%% file: v2.tex
\documentclass[11pt]{article}
\pdfoutput=1

\usepackage[final]{acl}

\usepackage{times}
\usepackage{latexsym}

\usepackage[T1]{fontenc}

\usepackage[utf8]{inputenc}

\usepackage{microtype}

\usepackage{inconsolata}

\usepackage{graphicx}
\usepackage{amssymb}
\usepackage{booktabs}
\usepackage{enumitem}
\usepackage{pifont}
\usepackage{url}

\title{EgoArgus: Benchmarking VLMs as Situational Assistants for Modality-Grounded User Supports}

\author{Yu-Chien Tang, Yu-Hsiang Liu, An-Zi Yen\\
       Department of Computer Science, National Yang Ming Chiao Tung University, Taiwan \\
      \texttt{tommytyc.cs10@nycu.edu.tw},
      \texttt{ivesliu.ee10@nycu.edu.tw},
      \texttt{azyen@nycu.edu.tw}}

\newcommand{\bench}{EgoArgus}

\newcommand{\vlms}{VLMs}

\begin{document}
\maketitle
\begin{abstract}
VLMs are increasingly positioned as daily assistants that perceive first-person environments, follow user dialogue, and decide how to help.
Existing egocentric benchmarks mainly evaluate visual understanding in isolation, leaving open whether models can arbitrate between visual evidence and user-provided language when the two are helpful, irrelevant, or conflicting.
We introduce \bench{}, a human-annotated dataset for evaluating egocentric assistants on understanding and decision tasks in five dialogue-video daily scenarios.
Our results demonstrate that it is still challenging for current VLMs as reliable egocentric assistants, which requires identifying which modality is trustworthy and deciding when intervention is warranted.
Deeper analysis also shows that existing modality bias mitigation methods are quite restricted to enhance performance, providing insights to aid practitioners into the deployment of current VLMs as daily assistants.\footnote{Our dataset is publicly available at \url{https://github.com/NYCU-NLP-Lab/EgoArgus}}
\end{abstract}

\section{Introduction}

Recent progress in vision-language models (\vlms) and multimodal large language models (MLLMs) has moved multimodal systems beyond passive image or video recognition toward interactive assistants that can perceive the environment, follow natural-language instructions, and answer user questions \citep{chen2026eagle,qwen3.5}.
This shift is especially important in egocentric settings, where wearable cameras or first-person robots observe daily activities from the user's point of view.
Several benchmarking studies argue that egocentric evaluation must move beyond narrow daily-life visual QA and cover broader domains or reasoning demands \citep{li2026egocross,pei2026egothinker}.

\begin{figure*}[t]
    \centering
    \includegraphics[width=\linewidth]{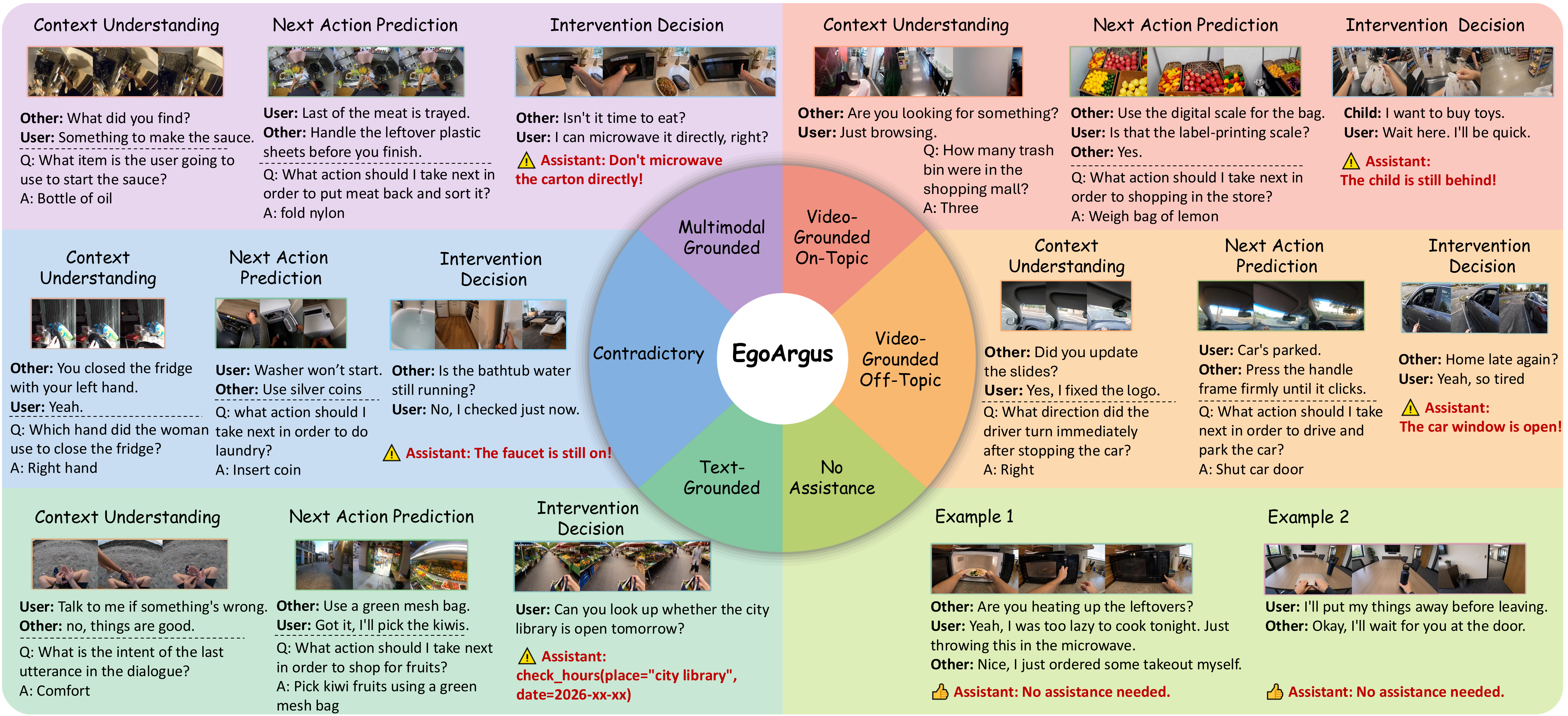} 
    \caption{Figure 1: Overview of \bench{}. \bench{} covers five assistance scenarios (multimodal grounded, contradictory, video-grounded on-topic, video-grounded off-topic, text-grounded) and a no-assistance case, each evaluated across three capabilities: context understanding, next action prediction, and intervention decision.}
    \label{fig:egoargus}
\end{figure*}

Recent datasets and benchmarks have begun to push first-person video evaluation toward assistant-like settings.
HoloAssist studies real instructor--performer collaboration with verbal guidance, mistake correction, and intervention behavior~\citep{wang2023holoassist}, while Ego-EXTRA emphasizes expert--trainee dialogue and VQA from the trainee's egocentric point of view~\citep{ragusa2026egoextra}.
Other work studies proactive assistant dialogue generation from streaming egocentric videos~\citep{zhang2025proassist}.
In parallel, streaming and proactive video benchmarks evaluate whether models can understand events online, wait for sufficient evidence, or respond at appropriate moments during an unfolding stream~\citep{lin2024streamingbench,niu2025ovobench,wang2025proactivevideoqa,zhang2025eyeswideopen}.
Together, these efforts mark a shift from static video understanding toward interactive first-person assistance.

To serve as a reliable daily assistant, a VLM must integrate first-person video with user dialogue and exhibit three progressively demanding capabilities: understanding the current situation, anticipating the action that should follow within an ongoing task, and proactively deciding whether, when, and how to intervene. 
We refer to these as context understanding, next action prediction, and intervention decision, respectively.
Taking the top-left group in Figure~\ref{fig:egoargus} as an example, context understanding requires the model to combine the user intent conveyed by the dialogue with the objects observed in the video to infer which item the user will use to start the sauce (a bottle of oil). 
Next action prediction requires reasoning over the dialogue instruction and the current scene to anticipate the action that should follow (folding the leftover plastic sheets). 
Intervention decision requires recognizing an impending hazardous action and deciding, without hesitation, when and what warning to raise (not to microwave the carton directly).

In real deployment, however, an assistant operates in a noisy environment. 
A VLM- or MLLM-based assistant continuously receives two streams at once: real-time first-person video and the linguistic input from user and surrounding dialogue. 
These two modalities are not necessarily consistent: they may complement one another, be mutually irrelevant, or directly conflict. 
The model therefore cannot treat all inputs as equally reliable. 
It must judge which signals to trust and which to disregard in the current situation, read the user's actual need, and, grounded in its understanding of the environment, offer an appropriate judgment and suggestion.

Studying how models judge, assist, and fail requires appropriate data.
Existing egocentric resources have advanced first-person video understanding, grounded QA, planning, and cross-domain QA~\citep{grauman2021ego4d,di2023grounded,qiu2024egoplanbench2,li2026egocross}.
Despite these advances, there has been limited exploration of controlled video--dialogue relations, particularly when paired with assist/no-assist labels, intervention timing, and assist-step annotations.
To this end, we build \bench{}, a resource for evaluating first-person assistants. 
As shown in Figure~\ref{fig:egoargus}, \bench{} is organized around five assistance scenarios: multimodal grounded, contradictory, video-grounded off-topic, video-grounded on-topic, and text-grounded. 
These scenarios cover the cases where dialogue is complementary to, conflicting with, irrelevant to, or itself the modality to rely on relative to the video. 
Beyond these, we additionally include a no-assistance case.
This reflects real deployment, where most situations do not call for intervention and an over-eager assistant becomes a distraction.

We construct \bench{} as two parts that together cover the three capabilities. 
The understanding part is a controlled MCQA suite of 6,978 examples from real egocentric daily-task videos~\citep{qiu2024egoplanbench2,di2023grounded}, covering context understanding and next-action prediction, paired with user dialogue~\citep{zhang2022mintrec} to instantiate the five scenarios. 
For the decision part, we use VISTA~\citep{liu2026vista} to synthesize 789 first-person assistant episodes that target intervention decision, difficult to collect densely in the real world, spanning the same scenarios and covering safety, non-safety, and no-assistance situations. 

We benchmark a range of recent VLMs on \bench{} and find a consistent text-dominant failure: when user dialogue contradicts what the camera shows, models follow the words rather than the scene, even though the scene plainly contains what they need. 
For context understanding, the assistant then reports a world that does not match reality, echoing the user's mistaken claim instead of correcting it; for next action prediction, this misreading propagates into guidance, steering the user toward a step that does not fit the actual scene. 
On the assistance episodes synthesized with VISTA, the dominant failure is the intervention decision itself: Models over-assist when they should stay quiet, and even when intervention is warranted, they misjudge its timing.
They may warn too early or too late, making the warning itself incorrect. 

\begin{figure*}[t]
  \centering
  \includegraphics[width=\textwidth]{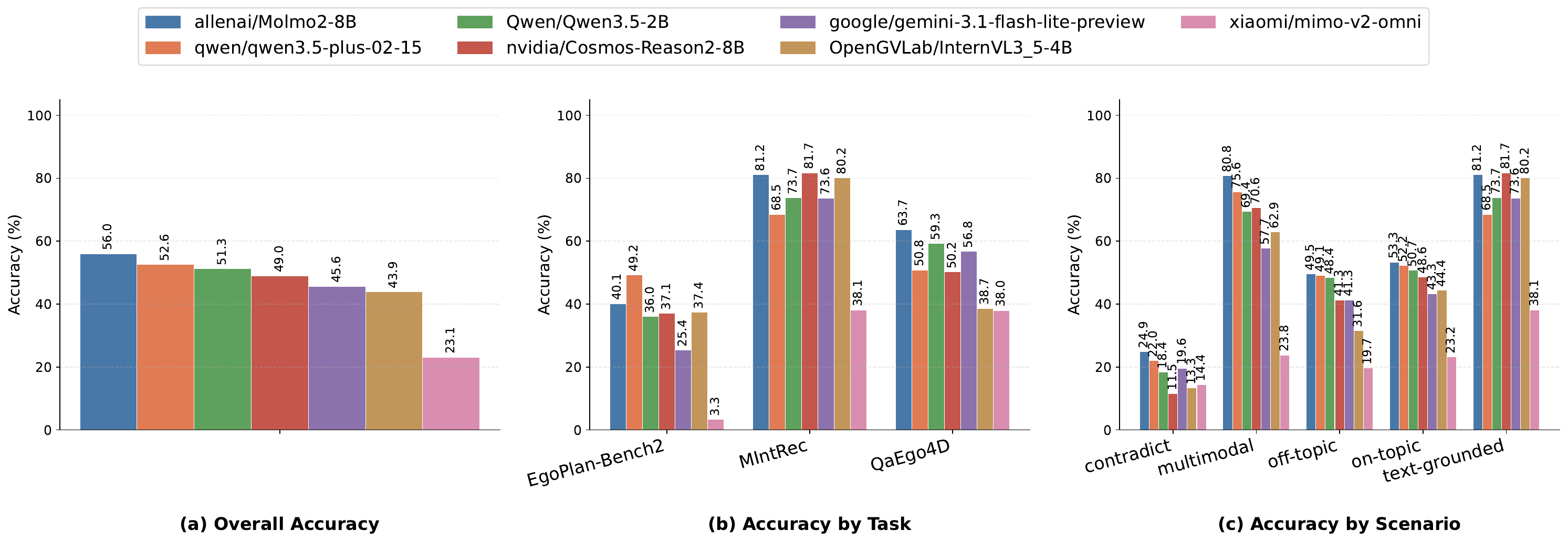}
  \caption{Results on understanding part of \bench{}. (a) The overall accuracy of each VLM. (b) The results divided by different tasks. (c) The accuracy of VLMs on the proposed 5 scenarios.}
  \label{fig:model-comparison}
\end{figure*}

In sum, our contributions can be summarized as follows:
(1) We introduce \bench{}, a human-annotated benchmark for evaluating egocentric assistants across MCQA and assist-step formats.
(2) We define five dialogue-video relations that cover multimodal grounded, contradictory, video grounded on/off topic, and text-grounded user dialogue, with additional labels for safety, non-safety, and no-assistance cases.
(3) We benchmark current VLMs and show large reliability gaps: contradictory dialogue drives real-video accuracy below chance, and accurately identifying the timing to provide assist steps are still challenging.
Deeper analysis further shows that existing modality bias mitigation methods provide limited enhancement, offering reflection on the real-world deployment of contemporary VLMs as daily assistant.

\section{\bench{} Construction}

\subsection{Resource Scope}
In EgoArgus, the understanding part combines three sources: EgoPlan-Bench2~\citep{qiu2024egoplanbench2} for next-action questions, QaEgo4D~\citep{di2023grounded,grauman2021ego4d} for grounded question answering over long egocentric videos, and MIntRec~\citep{zhang2022mintrec} for intent-labeled dialogue, which we expand into histories and reuse as off-topic distractors. 
The decision part is generated with VISTA~\citep{liu2026vista}, whose oracle metadata records whether help is required, the earliest useful intervention time, the issue to notice, and the recommended assistance step.
Table~\ref{tab:dataset} summarizes the data distribution.

\begin{table}[t]
  \centering
  \small
  \resizebox{\linewidth}{!}{\input{tables/dataset_composition.tex}}
  \caption{\bench{} composition in the current evaluated split. EgoPlan-Bench2, QaEgo4D, and MIntRec form the real-video MCQA part; VISTA contributes 789 evaluated Assist-Step rows across the same five scenarios plus no-assistance cases.}
  \label{tab:dataset}
\end{table}

\subsection{Five Assistance Scenarios}

The five scenarios are defined below, with intervention-decision examples from Figure~\ref{fig:egoargus}: 
\begin{itemize}[leftmargin=*,nosep] 
\item \textbf{Multimodal grounded}: the dialogue supplies an attribute, instruction, or relational context, while the video grounds the referenced object, action, or scene state. A correct judgment requires combining both. For example, warning against microwaving follows only from hearing the user's question and seeing that the container is a carton. 
\item \textbf{Contradictory}: the video determines the correct judgment, while the dialogue points to an incorrect one. For example, the user claims to have turned off the faucet, but the video shows it still running.  
\item \textbf{Video-grounded on-topic}: the video determines the correct judgment, while the dialogue is topically related but uninformative. For example, the user asks the child to wait and completes checkout before leaving, while the video shows that the child has been left behind. 
\item \textbf{Video-grounded off-topic}: the video determines the correct judgment, while the dialogue is unrelated. For example, the user makes idle talk about being home late, while the video shows the car window left open.
\item \textbf{Text-grounded}: the dialogue determines the correct judgment, while the video is irrelevant. For example, the user asks to check the library's opening hours, independent of the video. 
\end{itemize}

\subsection{Dialogue Construction and Verification}

We construct all examples by preserving the original answer-bearing source and varying only the relation between the user dialogue and the video question.
For the understanding part, we adopt EgoPlan-Bench2, QaEgo4D, and MIntRec for video and MCQA, and the addded dialogue is generated by prompting Gemini3.1-Pro across each scenario to be either misleading, unrelated, or merely topic-related. 
For the decision part, we leverage VISTA \citep{liu2026vista} to render a variety of scripts containing each scenario in video and prompt Gemini3.1-Pro to generate the dialogue accordingly.
Detailed real-video construction prompts, filtering rules, and aggregation procedures are provided in Appendix~\ref{app:construction} and~\ref{app:vista}.

\subsection{Data Annotation}
To verify the synthesized data, we invite 16 human annotators to check for both understanding and decision part. 
Each example is independently assigned to three annotators, and each annotator is asked to judge whether the constructed example satisfies its intended scenario and provide a corrected answer when the original gold option is invalid, ambiguous, or not supported by the video-dialogue pair.
Consequently, the verified answers are aggregated by majority vote and are used in the evaluated benchmark.

For the decision part, VISTA produces a larger candidate pool before filtering.
Reviewers discard generations whose rendering is physically implausible or mismatched to the intended scenario.
The evaluated split retains 311 unique reviewed videos, which are paired with scenario-specific dialogues to form 789 evaluated video-dialogue rows.
For assistance cases, the reviewed metadata specifies the appropriate intervention time and the help the assistant should provide.
For the no-assistance case, where the video depicts an ordinary daily activity, reviewers instead verify that the dialogue contains nothing that calls for assistance.

\section{Experiments}
\subsection{Experimental Setup}
\paragraph{Models}
For both parts of \bench{}, we evaluate seven VLMs: Molmo2-8B \citep{clark2026molmo2}, Qwen3.5-2B \citep{qwen3.5}, InternVL3.5-4B \citep{wang2025internvl35}, Gemini-3.1-Flash-Lite,\footnote{\url{https://blog.google/innovation-and-ai/models-and-research/gemini-models/gemini-3-1-flash-lite/}} Cosmos-Reason2-8B,\footnote{\url{https://huggingface.co/nvidia/Cosmos-Reason2-8B}} Qwen3.5-Plus \citep{qwen3.5}, and MiMo-V2-Omni.\footnote{\url{https://mimo.xiaomi.com/mimo-v2-omni}}

\paragraph{Metrics}
For the understanding part, we employ MCQA accuracy as the main evaluation metric.
To further observe the answering dynamics, we also report per-scenario accuracy to uncover the failure modes behind accuracy aggregation.
For the decision part, the model receives only the video, dialogue, and dialogue timestamp.
We evaluate whether the model correctly decides if assistance is required, and for required-assistance cases we also analyze the proposed intervention time, issue summary, evidence, and assist steps.
We report assistance-decision F1 and timing error when available, and use an LLM-assisted, frame-verified error taxonomy to separate context-understanding errors, next-action or assist-step planning errors, assistance-decision errors, and output-format failures.

\begin{table*}[t]
  \centering
  \small
  \resizebox{\textwidth}{!}{\input{tables/vista_model_error_table.tex}}
  \caption{Model performance and error analysis on VISTA. Rows are sorted by overall assistance-decision F1. We report overall F1, timing MAE, signed timing error $\Delta t$ (s), scenario-level F1 across the six VISTA scenarios, and mode-specific errors. Safety/non-safety FN are missed interventions; no-assist FP is false-alarm rate. Bold marks each model's strongest scenario score or lowest mode-specific error.}
  \label{tab:vista-model-error}
\end{table*}

\subsection{Results on Understanding Part}

The results are summarized in Figure~\ref{fig:model-comparison}, showing three main findings.

\noindent \textbf{Contradictory dialogue causes systematic failure.}
Mean accuracy falls from 63.0\% on multimodal-grounded examples to 18.5\% on contradictory examples.
Every evaluated model is below random guessing in the contradictory scenario except Molmo2-8B, which reaches 24.9\% and remains effectively at chance.
This result shows that the dialogue lure is not merely distracting; it actively reverses the decision in many cases.

\noindent \textbf{Text dominance is asymmetric.}
Mean text-grounded accuracy is 71.0\%, much higher than the video-grounded distractor scenarios.
This means models can often recover answers from dialogue even with irrelevant visual input, but they struggle to recover answers from video when dialogue provides a conflicting alternative.
The pattern is consistent with cross-modal imbalance reported in recent multimodal reasoning work \citep{wu2025mitigating,mullick-etal-2025-text}.

\noindent \textbf{Not all dialogue distractors are equally harmful.}
Off-topic dialogue reduces accuracy relative to multimodal examples, but its mean accuracy is still 41.5\%.
On-topic but answerless dialogue is slightly easier at 45.2\%.
The difference suggests that models can ignore clearly unrelated text better than actively contradictory text, and that some answerless on-topic dialogue may provide useful scene priors.

\subsection{Results on Decision Part}

The decision part evaluates whether the same modality-arbitration problem carries over from answer selection to assistant intervention.
Each synthetic episode contains a first-person generated video, dialogue context, and oracle assistance metadata, but models receive only the video, dialogue, and dialogue timestamp.
The model must decide whether assistance is required and, when it is, produce an intervention time, issue summary, supporting evidence, and assist steps.
Table~\ref{tab:vista-model-error} shows that decision performance varies widely across models.

\noindent \textbf{Decision ability differs sharply by model.}
The strongest models make more reliable intervention decisions, while smaller or weaker models often fail even with full-video input.
This trend suggests that stronger instruction-following helps, but does not remove the core decision difficulty.

\noindent \textbf{No-assistance cases remain difficult.}
Models that perform well on assistance-required scenarios still struggle to remain silent when the scene is benign.
This failure matters for deployment because an overactive assistant can become disruptive even when it understands parts of the scene.

\noindent \textbf{Intervention policy shows a precision--recall trade-off.}
Some models catch more safety and non-safety assistance needs but also raise more false alarms, while more conservative models reduce false alarms at the cost of missing required interventions.
Simple conservatism is therefore not a usable assistant policy.

\noindent \textbf{Timing is systematically imperfect.}
VLMs tend to intervene before the reviewed oracle timing, and their timing errors remain large enough to affect short safety or task-assistance events.
The decision part therefore requires calibrated intervention timing, not only recognizing that assistance may be relevant.

The post-hoc audit further shows that assistance-decision errors, assist-step planning errors, and context-understanding errors all contribute substantially to model failures.
This distribution complements the understanding part results: many failures occur not only when a model cannot perceive the relevant context, but also when it must decide whether the perceived situation needs intervention.

\begin{figure*}[t]
  \centering
  \includegraphics[width=\textwidth]{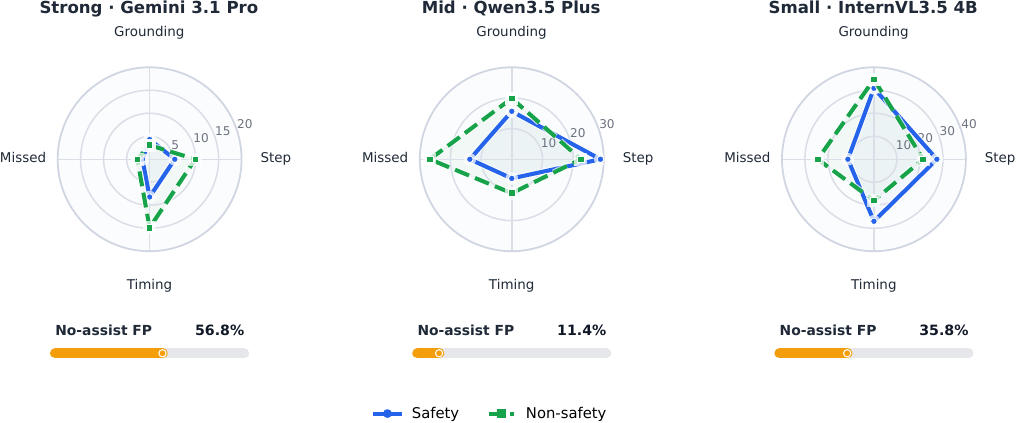}
  \caption{Representative model error signatures on intervention decision. We additionally evaluate Gemini-3.1-Pro as the strong-model reference.}
  \label{fig:vista-error-signature-radar}
\end{figure*}

\subsection{Intervention Failures}

We group intervention decision errors into five types. 
Grounding errors occur when the model fails to understand the relevant evidence in the scene or dialogue. 
Step errors occur when the model recognizes that assistance is needed but proposes an inadequate assist step that is vague, incomplete, or misaligned with the oracle response. 
Timing errors occur when the model intervenes at the wrong moment, either after the event has already happened or before sufficient evidence is available. 
Missed interventions are cases where the oracle marks assistance as necessary but the model does not intervene, similar to false negatives. 
No-assist false alarms (FP) are the opposite, where the oracle requires no assistance but the model intervenes anyway.

As shown in Figure~\ref{fig:vista-error-signature-radar}, 
the strong model (Gemini-3.1-Pro) makes the fewest errors overall, but at the cost of the highest no-assist false-alarm rate, over-intervening in benign scenes. 
The mid model (Qwen3.5-Plus) keeps false alarms low and timing relatively stable, but its errors concentrate on missed interventions, especially for non-safety cases. 
The small model (InternVL3.5-4B) has the largest errors overall. 
Beyond the worst timing, which is most severe on safety cases, it also makes frequent step errors.
This is a deployment concern, since models at the 4B scale are the most realistic candidates for on-device wearable use.

\section{Discussion}
\subsection{Impacts of Distractor Modality}
To empirically validate whether the low accuracy comes from the distractor modality, we run an oracle diagnostic on Qwen3.5-2B, InternVL3.5-4B, Cosmos-Reason2-8B, and Molmo2-8B using the same 1,000 balanced examples (200 per scenario).
At test time, we remove dialogue for contradictory, off-topic, and on-topic video-grounded examples, and remove video for text-grounded examples.
The multimodal-grounded setting is unchanged.

\begin{table*}[t]
  \centering
  \small
  \resizebox{\textwidth}{!}{\input{tables/drop_distractor.tex}}
  \caption{Full-input and oracle distractor-removal accuracy (\%) for four VLMs on the same 1,000 balanced examples.}
  \label{tab:drop}
\end{table*}

Table~\ref{tab:drop} shows a consistent and substantial improvement in the contradictory setting.
Removing contradictory dialogue improves accuracy by 34.0--43.0 points across all four models, confirming that contradictory dialogue is the primary driver of the severe video-grounded failure.
In contrast, off-topic and text-grounded changes stay within 2.0 points, while removing on-topic dialogue is neutral to detrimental (up to $-4.5$ points).
Thus, a useful mitigation cannot simply delete or downweight all non-answer modalities: answerless but on-topic dialogue can still provide context, and text-grounded examples require preserving the dialogue channel.

\subsection{Modality Representation Separation}

\begin{table}[t]
  \centering
  {\footnotesize
  \setlength{\tabcolsep}{3pt}
  \input{tables/linear_probe_summary.tex}}
  \caption{Cross-model layer-wise linear-probe summary for four VLMs. Jump reports the layer and fractional decoder depth; pre, post, and peak values are probe accuracies (\%).}
  \label{tab:linear-probe}
\end{table}

We further ask whether failures arise because the model's internal representations do not clearly separate the video and dialogue sources at answer time.
Following layer-wise probing analyses of modality preference \citep{yan2026beyond}, we train a separate linear probe for each decoder layer of Qwen3.5-2B, InternVL3.5-4B, Cosmos-Reason2-8B, and Molmo2-8B.
For each example, the input to the probe is the last-token hidden state at that layer, and the target is the model's own final soft distribution over the four answer options.
This setup diagnoses where the model's answer preference becomes linearly recoverable; it does not use gold correctness as the probe target.

\begin{figure}[t]
  \centering
  \includegraphics[width=\linewidth]{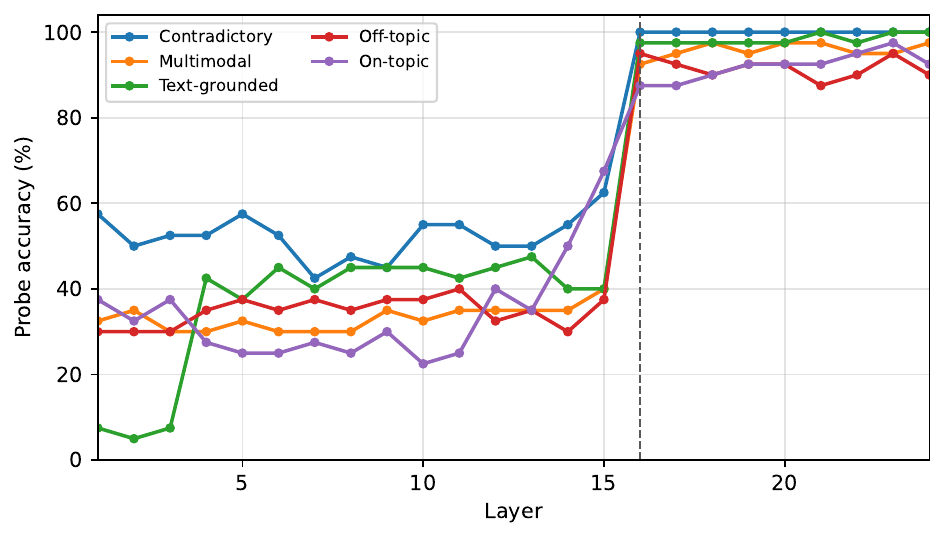}
  \caption{Detailed layer-wise probing diagnostics for Qwen3.5-2B. Answer-preference information is weakly decodable through layer 15, then jumps sharply at layer 16 across all five scenarios.}
  \label{fig:linear-probe-accuracy}
\end{figure}

Table~\ref{tab:linear-probe} shows the same abrupt transition across all four models.
Qwen3.5-2B jumps from 49.5\% at layer 15 to 94.5\% at layer 16, while the three 36-layer models jump at layer 25 from 60.0--69.5\% to 93.5--97.0\%.
The transition consistently occurs at middle to deep fractional decoder depth, and peak probe accuracy reaches 97.5--100.0\% across models.
Figure~\ref{fig:linear-probe-accuracy} provides a more detailed Qwen3.5-2B view across each layers.
It can be seen that the pattern appears across all five scenarios: for example, text-grounded examples rise from 40.0\% at layer 15 to 97.5\% at layer 16, and contradictory examples reach 100.0\% at layer 16.

\begin{figure}[t]
  \centering
  \includegraphics[width=\linewidth]{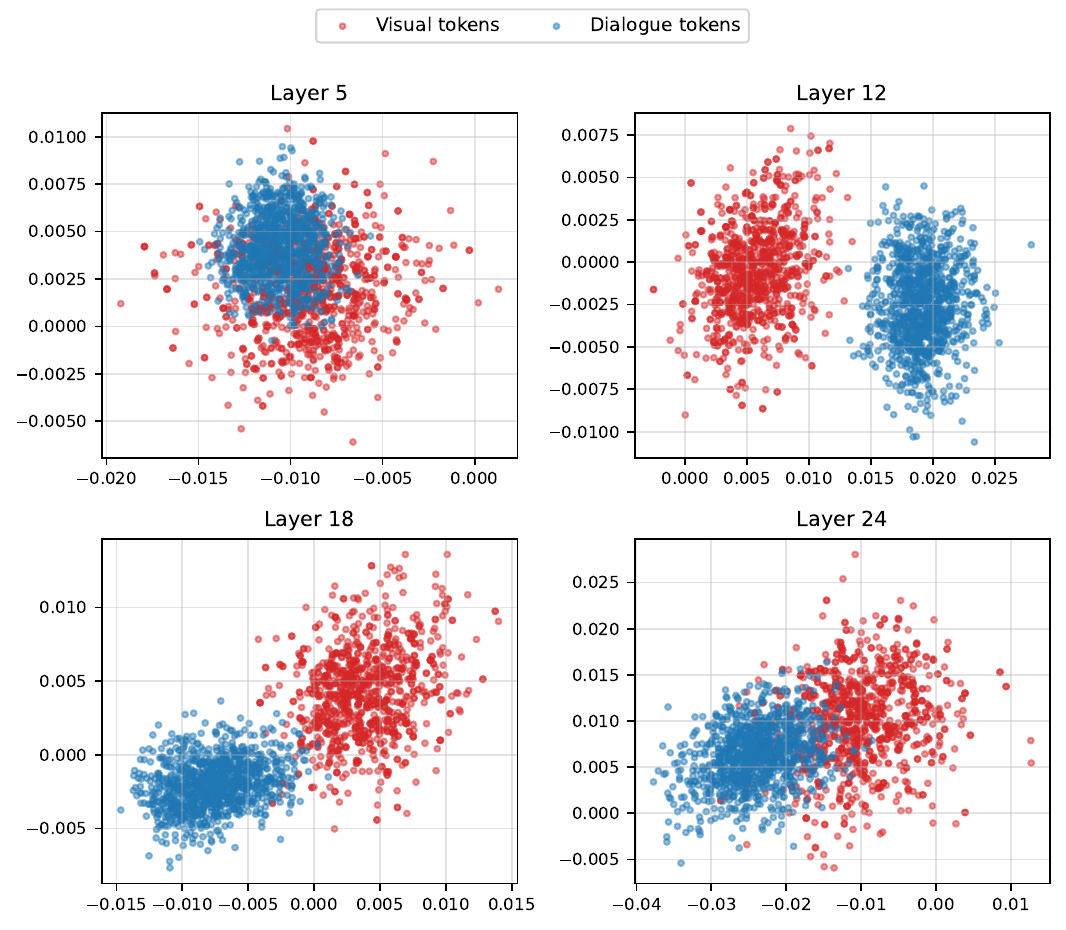}
  \caption{Qwen3.5-2B visual-token and dialogue-token representations in the linear-probe weight space. The two modalities are mixed in shallow layers and become separated in middle-to-late layers.}
  \label{fig:linear-probe-svd}
\end{figure}

Figure~\ref{fig:linear-probe-svd} gives a complementary Qwen3.5-2B view of token representations.
Visual and dialogue token averages are heavily mixed at layer 5, become visibly separated by layers 12 and 18, and remain mostly separated at layer 24.
Together, the cross-model probes show that late answer-preference formation is a general phenomenon for VLMs, while the token visualization shows that separating visual and textual sources does not guarantee correct modality arbitration: a model may distinguish the two sources yet still follow misleading dialogue when deciding the answer.

\subsection{Modality Bias Mitigation}

\begin{table}[t]
  \centering
  \small
  \resizebox{\linewidth}{!}{\input{tables/uniform_attention.tex}}
  \caption{Results under uniform visual-token attention reweighting. The intervention is sensitive to $\epsilon$ and does not uniformly improve all scenarios.}
  \label{tab:uniform}
\end{table}

To delve deeper into whether existing modality bias mitigation methods can effectively enhance the performance, we conduct experiment with training-free and training-based baselines.
\subsubsection{Uniform Attention Reweighting}
We implement an inference-time attention manipulation baseline following the general idea of modal reweighting \citep{wu2025mitigating}.
During decoding, the intervention adds a scalar $\epsilon$ to attention logits for visual tokens before the softmax:
$\mathrm{softmax}(\log w_t + \epsilon \mathbf{1}_{C_v})$,
where $C_v$ indexes visual tokens.
Positive $\epsilon$ increases visual attention, while negative $\epsilon$ decreases it.
We evaluate Qwen3.5-2B and report the accuracy over the five scenarios in Table~\ref{tab:uniform}.
The results show that uniform reweighting is brittle.
Moderate interventions leave several scenarios near baseline, but strong positive reweighting damages multimodal, off-topic, on-topic, and text-grounded performance.
The text-grounded degradation is expected: when the answer is in dialogue, indiscriminately increasing visual attention amplifies an irrelevant modality.
More importantly, uniform reweighting does not reliably solve the impact of contradictory dialogue, suggesting that increasing visual attention alone may be insufficient once the language context strongly lures the model.

\subsubsection{Preference Alignment Pipeline}

\begin{table}[t]
  \centering
  \small
  \input{tables/napo_alignment.tex}
  \caption{Results before and after NaPO-style preference alignment on the understanding part real-video MCQA.}
  \label{tab:napo}
\end{table}

We also implement NaPO, a training-time route based on noise-aware preference optimization \citep{zhang2025napo}.
The pipeline constructs general, language-biased, and vision-biased preference pairs from an external image-text preference dataset, trains a Qwen3.5-2B LoRA adapter with the NaPO objective, and evaluates the adapted model on the five real-video MCQA scenarios in the understanding part.
We follow the NaPO paper and employ RLAIF-V dataset \citep{Yu_2025_CVPR} to construct the preference pairs, and the results are summarized in Table~\ref{tab:napo}.
It can be seen that this generic preference-alignment route does not resolve dialogue-induced modality bias.
NaPO slightly improves text-grounded accuracy, where the answer-bearing evidence is linguistic, but reduces accuracy on multimodal, contradictory, off-topic, and on-topic examples.
Overall accuracy drops from 51.3\% to 48.9\%.
This pattern suggests that preference alignment derived from image-text preference pairs may improve language-following behavior without teaching the assistant when to override, ignore, or reconcile user dialogue against egocentric video evidence.

\section{Related Work}

\paragraph{Egocentric video understanding.}
Ego4D established a large-scale foundation for first-person video understanding \citep{grauman2021ego4d}.
EgoPlan-Bench2 evaluates real-world planning in egocentric videos \citep{qiu2024egoplanbench2}, while QaEgo4D studies grounded question answering over long egocentric videos \citep{di2023grounded}.
Recent resource benchmarks broaden egocentric evaluation beyond standard daily-life QA; for example, EgoCross evaluates cross-domain generalization across surgery, industry, extreme sports, and animal-perspective videos \citep{li2026egocross}.
\bench{} follows this resource-paper motivation but targets a different deployment gap: whether VLMs can serve as assistants when first-person video must be interpreted together with user dialogue.
It reuses EgoPlan-Bench2 and QaEgo4D as real-video sources but changes the evaluation target from visual understanding alone to modality arbitration and assistance reliability.
We further use VISTA \citep{liu2026vista} to construct controlled first-person assistant episodes, extending the benchmark from real-video answer selection to synthetic intervention decisions.

\paragraph{Multimodal conversation and intent.}
MIntRec provides large-scale multimodal conversational intent annotations \citep{zhang2022mintrec}.
We use its dialogue content to create both off-topic distractors and text-grounded examples.
This lets the benchmark include natural conversational language rather than relying only on synthetic distractor text.

\paragraph{Modality bias and mitigation.}
Recent work shows that multimodal models may exhibit cross-modal attention imbalance and can fail to reconcile conflicting evidence across modalities \citep{wu2025mitigating}.
Recent probing work further studies how modality preference can be localized inside omni-modal models through modality selection metrics and layer-wise linear probes \citep{yan2026beyond}.
Preference optimization has also been proposed as a way to reduce modality bias and hallucination \citep{zhang2025napo}.
VEA studies the gap between attending to visual evidence and using that evidence in the final answer \citep{liu2025seeing}.
Our work complements these efforts by placing modality bias in an egocentric assistant setting where user dialogue can help, distract, or contradict the video.

\section{Conclusion}

We introduced \bench, a two-part benchmark for dialogue-induced modality bias in egocentric video assistants.
Across seven VLMs, contradictory dialogue causes the strongest failure, reducing mean accuracy below random guessing despite the answer being visible in video.
The VISTA part extends this diagnosis to synthetic assistant episodes, where deciding whether and how to intervene is a central audited failure layer.
Oracle distractor removal confirms that modality interference drives many errors, and linear probing shows that answer preferences become recoverable only after late-middle layers where visual and dialogue tokens separate.
Attention-based mitigation further shows that fixed visual reweighting is not enough.
Reliable egocentric assistants need adaptive mechanisms that infer which modality is answer-bearing, which is a distractor, and when apparently non-answer context should still be preserved.

\section*{Limitations}

\bench{} has two evaluation formats, and each has different limitations.
The understanding part focuses on four-choice VQA, which simplifies answer extraction and makes random-choice behavior interpretable but does not cover open-ended assistant behavior.
The decision part addresses open assistant intervention, but its videos are synthetic and may not capture the full visual diversity, sensor noise, and social ambiguity of deployed egocentric assistants.
The current evaluated split contains 6,978 real-video MCQA examples and 789 VISTA Assist-Step rows; larger candidate pools and future VISTA generations are not all part of the reported evaluation.
The layer-wise probing analysis covers Qwen3.5-2B, InternVL3.5-4B, Cosmos-Reason2-8B, and Molmo2-8B, while the attention-intervention and NaPO analyses remain specific to Qwen3.5-2B because they require model-specific attention internals or training.
Extending the same intervention analysis to all evaluated models remains in our future work.

\section*{Ethics Statement}

The real-video part uses egocentric video sources whose original releases include privacy and consent procedures.
The VISTA part uses generated first-person videos and human-reviewed synthetic scenarios, which reduces direct privacy exposure but still requires care because the episodes may depict safety-relevant assistance situations.
Our added dialogues can intentionally contradict the visual evidence, so released examples should be clearly labeled as diagnostic, synthetic, or constructed contexts rather than treated as faithful transcripts.
The goal is to evaluate assistant reliability under misleading user context, not to model or profile real user behavior.
For deployed assistants, the results argue against blindly trusting any single modality, especially user-provided text that may be mistaken or adversarial.

\section*{Acknowledgments}
This research was partially supported by National Science and Technology Council, Taiwan, under grants NSTC 114-2221-E-A49-057-MY3, NSTC 114-2221-E-002-070-MY3, and NSTC 115-2634-F-002-012-, and Ministry of Education (MOE) in Taiwan, under grants 115L900901.

\bibliography{custom}

\appendix

\section{Understanding Part Construction Details}
\label{app:construction}

\subsection{Multimodal-Grounded Generation}

For EgoPlan-Bench2, we start from the original next-action MCQA examples.
The generator receives the original question, answer choices, gold answer, and available visual context, then writes a 4--5 turn user-assistant style dialogue.
The dialogue is required to provide a missing instruction, constraint, or method that helps determine the visually grounded next action, while it must not directly reveal the answer option.
The original video remains necessary because the dialogue describes what the user wants or how the action should be interpreted, not the full visual state.

For QaEgo4D, we use a stricter multi-step pipeline because the source questions cover diverse long-video QA phenomena.
The pipeline first selects a multimodal question pattern, then generates dialogue that contributes non-visual context, and finally audits the constructed MCQA.
During the audit step, examples are rejected or revised when the answer can be obtained from video alone, from dialogue alone, or from superficial wording in the choices.
This verification is intended to make the multimodal-grounded split test cooperative use of both modalities rather than ordinary video QA with extra text.

\subsection{Contradictory Generation}

Contradictory examples preserve the original video-grounded gold answer from EgoPlan-Bench2 or QaEgo4D.
The generator is instructed to create dialogue that is fluent and plausible but factually conflicts with the video evidence.
The misleading dialogue may describe an incorrect object, action, location, state, or temporal relation, and at least one distractor choice is aligned with this erroneous dialogue.
Thus, the model must rely on the video to answer correctly and must ignore the user dialogue when it conflicts with visual evidence.

\subsection{Distractor and Text-Grounded Scenarios}

For video-grounded off-topic examples, we keep the original EgoPlan-Bench2 and QaEgo4D video questions and randomly pair them with unrelated MIntRec dialogue histories.
For video-grounded on-topic examples, we generate dialogue from nearby narration events, video category information, or question context, but explicitly forbid mentioning the gold answer or directly answering the question.
The off-topic and on-topic subsets share the same sampled video-question identities, which isolates the effect of dialogue relevance while keeping the visual QA task fixed.

For text-grounded examples, MIntRec utterances are expanded into short dialogue histories ending in the original intent-bearing utterance.
The question asks for the intent of the final utterance, and the answer choices are intent labels.
Each text-grounded example is paired with a random egocentric video distractor, making the dialogue the only answer-bearing modality.

\subsection{Human Verification Workflow}

\begin{figure}
    \centering
    \includegraphics[width=\linewidth]{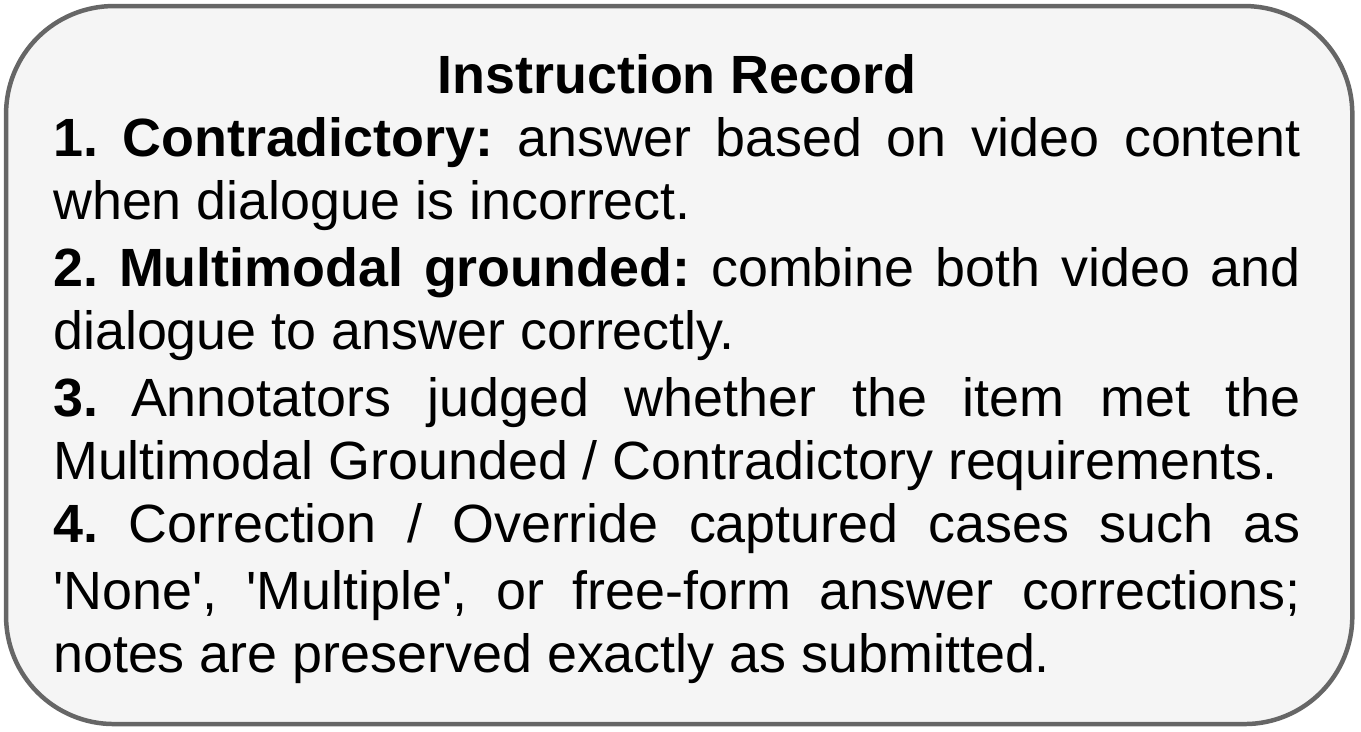}
    \caption{The annotation instructions for human annotators.}
    \label{fig:annotation_instruction}
\end{figure}

Human verification focuses on multimodal-grounded and contradictory examples because these scenarios depend on precise semantic relations between generated dialogue, video evidence, and answer choices.
Each candidate example is labeled by three annotators.
Annotators mark whether the example satisfies the target scenario, select or correct the answer, and can leave free-form comments for ambiguous or invalid cases.
We use majority voting over corrected answers to produce the final verified labels.
Examples that annotators identify as invalid or underspecified are resolved through correction or follow-up review before the final benchmark aggregation.
The full annotation instructions are shown in Figure \ref{fig:annotation_instruction}. 

\section{Decision Part Construction Details}
\label{app:vista}

\subsection{Synthetic Episode Construction}

The VISTA part is designed to cover daily-assistance cases that are difficult to collect densely with real egocentric video.
Starting from a reviewed seed, the construction pipeline specifies the intervention target, the user's visible action, the warning or assistance signal, and a structured render script.
The generated video is paired with a dialogue context and oracle metadata describing whether help is required, when the earliest useful intervention occurs, what issue should be noticed, and what step should be recommended.
This metadata is used for evaluation and analysis, but it is never provided to the model.

The exported VISTA rows span the same five dialogue-video relations as the real-video benchmark: multimodal grounded, contradictory, video-grounded off-topic, video-grounded on-topic, and text-grounded.
The synthetic part also includes no-assistance controls, where the correct assistant behavior is to avoid raising an intervention.
We further label assistance cases as safety assistance, non-safety assistance, or no assistance, which separates missed hazards from missed practical help and false alarms.

\begin{figure*}[p]
  \centering
  \includegraphics[width=\textwidth]{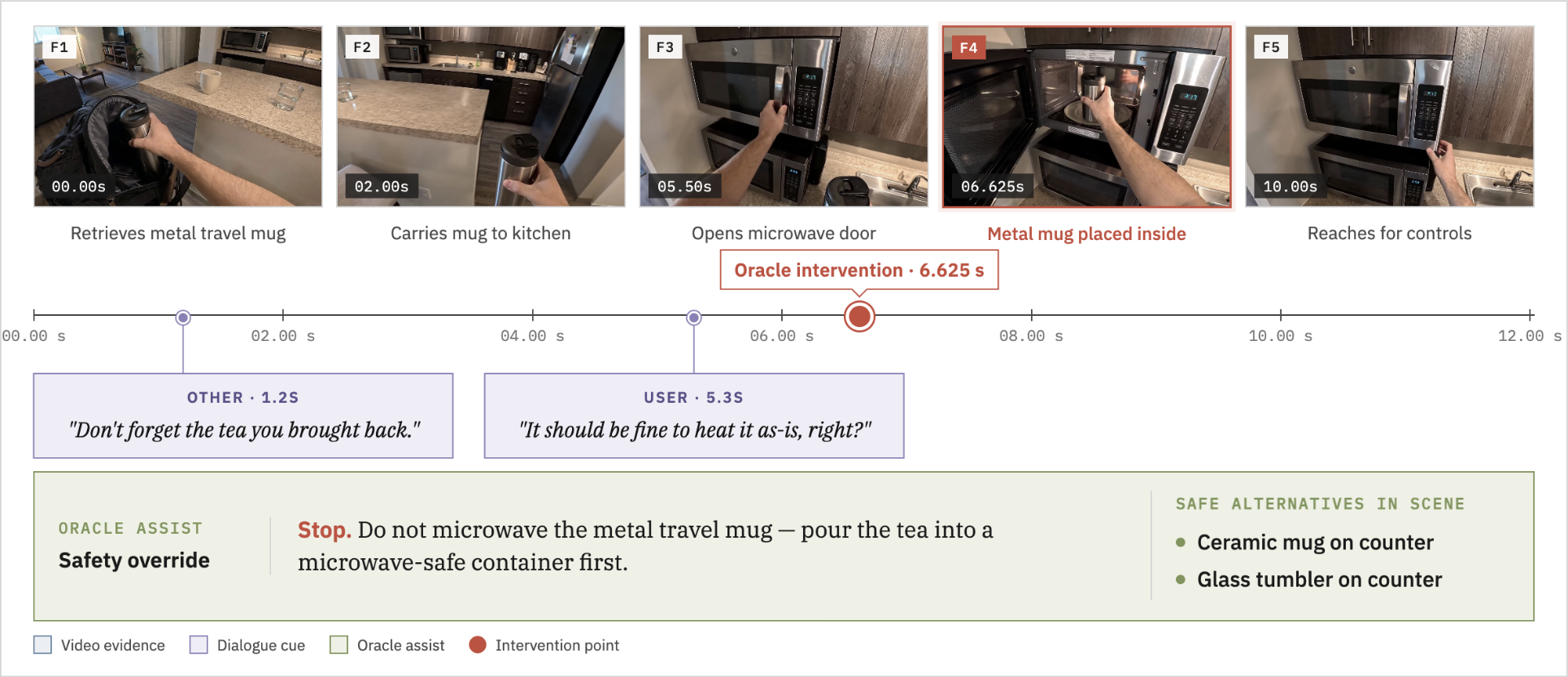}
  \caption{An annotated decision-part example from a safety assistance case. The VISTA-generated video (F1–F5) shows a user placing a metal travel mug into a microwave, while the dialogue gives no sign of any problem. Recognizing the hazard requires the video, since only the visual content reveals that the container is metal. Annotators label the earliest useful intervention time (6.625,s), the issue to warn about, and the assistance to provide, including the microwave-safe alternatives visible in the scene.}
  \label{fig:vista-microwave-showcase}
\end{figure*}

\subsection{Human-in-the-Loop Review}

Because VISTA videos are generated, quality control is part of dataset construction.
Reviewers inspect whether the intended event is visible, whether the first-person viewpoint remains plausible, whether object identity is stable, whether the dialogue relation matches the target scenario, and whether the video contains evidence for the oracle assistance label.
Failed generations are discarded or revised through script editing and regeneration.

The current human-in-the-loop audit trail contains 2,917 events across 218 cases.
These include 565 attempt-review events, 447 script-revision files, 204 regeneration attempts, and 1,466 human-authored events.
Common review tags include object-identity drift, assistance-signal problems, point-of-view breaks, spatial layout flips, seed mismatch, and no-assistance semantic errors.
The round-1 reviewed export contains 410 usable video-dialogue pairings after filtering for saved dialogue and reviewed video availability.

\subsection{Decision-Part Annotation}

Figure~\ref{fig:vista-microwave-showcase} shows an annotated decision-part example.
The VISTA-generated video depicts a user retrieving a metal travel mug, carrying it to the kitchen, and placing it inside a microwave, while the dialogue—an offhand remark followed by the user's own ``It should be fine to heat it as-is, right?''---gives no indication that anything is wrong. Recognizing the hazard requires combining both: only the video reveals that the container is metal. For each such case, annotators mark the earliest useful intervention time (here 6.625s, the moment the mug is placed inside), the issue to warn about, and the assistance the model should provide, namely stopping the user and pointing to the microwave-safe alternatives visible in the scene (a ceramic mug and a glass tumbler on the counter).

\subsection{Decision-Part Composition}

\begin{figure*}[t]
  \centering
  \includegraphics[width=\textwidth]{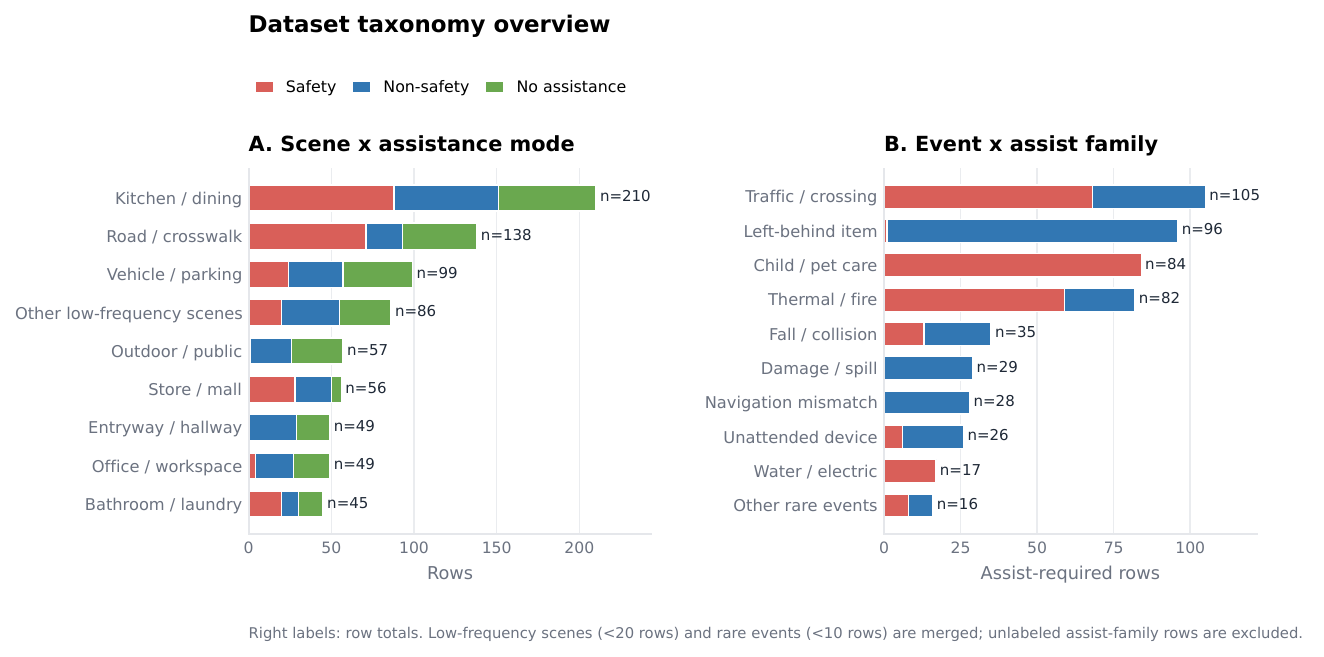}
  \caption{Composition of the decision part. (A) Examples by scene, stacked by assistance mode (safety, non-safety, no-assistance). (B) Assist-required examples by event type, split into safety and non-safety.}
  \label{fig:vista-taxonomy-overview}
\end{figure*}

Figure~\ref{fig:vista-taxonomy-overview} summarizes the distribution of the decision part along two axes: scene and event. 
Panel~A counts examples by scene, stacked by the three assistance modes (safety, non-safety, and no-assistance). 
The data spans diverse everyday scenes, led by kitchen/dining and road/crosswalk, and most scenes contain both assist-required and no-assistance cases, so that a model cannot infer whether to intervene from the scene type alone. 
Panel~B restricts to assist-required examples and counts them by assist family, separating safety from non-safety. 
Assist-required events are dominated by traffic/crossing, left-behind items, child/pet care, and thermal/fire; traffic, child care, and thermal events are predominantly safety, whereas left-behind items and navigation mismatches lean non-safety. 
Overall, the data is diverse across both scenes and events and mixes multiple assistance modes within each scene, allowing the decision part to evaluate intervention under a realistic distribution.

\subsection{Comparison with Related Benchmarks}

\input{tables/vista_related_work_matrix}

Table~\ref{tab:benchmark-comparison} positions EgoArgus against representative egocentric and assistant benchmarks along six axes central to our setting: first-person video, user dialogue, the intervention decision, intervention timing, assist steps, and no-assistance handling. 
Existing benchmarks tend to cover only a subset. 
One group pairs first-person video with dialogue but treats decision, timing, steps, and no-assistance only partially. 
A second group targets streaming or timing-aware evaluation yet largely omits dialogue and the decision of whether to act. 
A third group of proactive-assistant benchmarks covers decision and steps more directly, but few combine dialogue with explicit timing and no-assistance coverage. 
EgoArgus is the only resource that provides primary coverage on all six axes, jointly evaluating whether, when, and how to intervene, grounded in first-person video and user dialogue and including no-assistance cases where the correct behavior is to stay silent.

\section{Linear Probe Details}
\label{app:linear-probe}

For each of Qwen3.5-2B, InternVL3.5-4B, Cosmos-Reason2-8B, and Molmo2-8B, the probing analysis uses the same 1,000 real-video MCQA examples, with 200 examples sampled from each of the five scenarios.
For each example, we extract hidden states from all decoder layers: 24 layers for Qwen3.5-2B and 36 layers for each of the other models.
The main probe uses the hidden state at the final token position, after $L_2$ normalization, because this position aggregates the prompt context used for next-token answer prediction.
The probe target is a soft label formed by applying a softmax to the model's own logits for the four answer letters A--D.

For each layer, we train an independent linear classifier with soft cross-entropy loss.
The split is balanced by scenario, with 60\% for training, 20\% for validation, and 20\% for testing.
We train each probe with Adam, learning rate $10^{-3}$, batch size 256, and 200 epochs, selecting the checkpoint with the lowest validation loss.
For the SVD visualization, we average visual-token hidden states and dialogue-token hidden states separately at each layer, decompose the corresponding probe weight matrix, and project both token averages onto the top two right singular vectors.

\end{document}

%% file: tables/dataset_composition.tex
\begin{tabular}{lrrrrr}
\toprule
Scenario & EgoPlan & QaEgo4D & MIntRec & VISTA & Total \\
\midrule
Multimodal & 500 & 500 & 0 & 116 & 1,116 \\
Contradictory & 500 & 500 & 0 & 104 & 1,104 \\
Off-topic & 1,000 & 1,000 & 0 & 118 & 2,118 \\
On-topic & 1,000 & 1,000 & 0 & 102 & 2,102 \\
Text-grounded & 0 & 0 & 978 & 110 & 1,088 \\
No assistance & 0 & 0 & 0 & 239 & 239 \\
\midrule
Total & 3,000 & 3,000 & 978 & 789 & 7,767 \\
\bottomrule
\end{tabular}

%% file: tables/vista_model_error_table.tex
\begin{tabular}{@{}llrrrrrrrrrrrr@{}}
\toprule
\textbf{Model} &
\textbf{Input} &
\textbf{F1} $\uparrow$ &
\textbf{MAE} $\downarrow$ &
\textbf{$\Delta t$} &
\multicolumn{6}{c}{\textbf{Scenario F1} $\uparrow$} &
\multicolumn{3}{c}{\textbf{Mode error} $\downarrow$} \\
\cmidrule(lr){6-11}\cmidrule(l){12-14}
 & & & & &
Multi. &
Contra. &
Vid on &
Vid off &
Text &
No asst. &
Safety FN &
Non-saf. FN &
No-asst. FP \\
\midrule
Gemini 3.1 Flash Lite & Full-video & 0.8117 & 2.7420 & -2.4970 & 0.9395 & 0.9042 & 0.8049 & 0.7431 & \textbf{0.9463} & 0.3885 & \textbf{0.1445} & 0.2443 & 0.3173 \\
MiMo V2 Omni & Full-video & 0.7500 & 2.6340 & -2.3420 & \textbf{0.9194} & 0.8681 & 0.8171 & 0.7046 & 0.7031 & 0.3051 & \textbf{0.1992} & 0.3969 & 0.2620 \\
Qwen3.5 Plus 02-15 & Full-video & 0.7232 & 2.5050 & -1.9280 & \textbf{0.9340} & 0.9101 & 0.7249 & 0.6087 & 0.4444 & 0.4418 & 0.3008 & 0.4695 & \textbf{0.1144} \\
InternVL3.5 4B & Sampled & 0.5979 & 3.7030 & -0.4040 & \textbf{0.9091} & 0.8587 & 0.8571 & 0.1935 & 0.0909 & 0.2483 & 0.4141 & 0.5687 & \textbf{0.3506} \\
Cosmos Reason2 8B & Sampled & 0.5572 & 2.7840 & -1.7310 & 0.7283 & \textbf{0.7577} & 0.6131 & 0.5165 & 0.1897 & 0.3095 & 0.4023 & 0.7595 & \textbf{0.1070} \\
Molmo2 8B & Sampled & 0.5375 & 4.0190 & -3.9950 & 0.8265 & \textbf{0.8603} & 0.7655 & 0.0854 & 0.1251 & 0.1961 & 0.4922 & 0.6107 & \textbf{0.3985} \\
Qwen3.5 2B & Full-video & 0.1920 & 3.7840 & -3.7840 & \textbf{0.3459} & 0.1509 & 0.2095 & 0.0854 & 0.2203 & 0.0000 & 0.8438 & 0.9237 & \textbf{0.0037} \\
\bottomrule
\end{tabular}

%% file: tables/drop_distractor.tex
\begin{tabular}{llrrrrrr}
\toprule
Model & Input & Multimodal & Contradictory & Off-topic & On-topic & Text-grounded & Mean \\
\midrule
Qwen3.5-2B & Full & 64.0 & 16.0 & 48.5 & 50.5 & 73.5 & 50.5 \\
 & Drop distractor & 64.0 & 57.0 & 49.5 & 49.5 & 73.0 & 58.6 \\
\midrule
InternVL3.5-4B & Full & 73.5 & 13.0 & 50.0 & 49.5 & 76.0 & 52.4 \\
 & Drop distractor & 73.5 & 56.0 & 49.0 & 49.0 & 77.0 & 60.9 \\
\midrule
Cosmos-Reason2-8B & Full & 77.0 & 10.5 & 49.0 & 54.5 & 81.0 & 54.4 \\
 & Drop distractor & 77.0 & 51.5 & 50.0 & 50.0 & 82.5 & 62.2 \\
\midrule
Molmo2-8B & Full & 81.5 & 27.5 & 54.5 & 56.0 & 79.5 & 59.8 \\
 & Drop distractor & 81.5 & 61.5 & 56.5 & 56.5 & 81.0 & 67.4 \\
\bottomrule
\end{tabular}

%% file: tables/linear_probe_summary.tex
\begin{tabular}{lccccc}
\toprule
Model & \#L & Jump & Pre & Post & Peak \\
\midrule
Qwen3.5 & 24 & L16 (.67) & 49.5 & 94.5 & 97.5@L23 \\
InternVL3.5 & 36 & L25 (.69) & 60.0 & 96.0 & 99.0@L34 \\
Cosmos-R2 & 36 & L25 (.69) & 69.5 & 93.5 & 98.5@L31 \\
Molmo2 & 36 & L25 (.69) & 65.5 & 97.0 & 100.0@L31 \\
\bottomrule
\end{tabular}

%% file: tables/uniform_attention.tex
\begin{tabular}{lrrrr}
\toprule
Scenario & $\epsilon=-10$ & $\epsilon=-5$ & $\epsilon=5$ & $\epsilon=10$ \\
\midrule
Multimodal & 67.5 & 67.3 & 68.2 & 36.8 \\
Contradictory & 15.9 & 15.9 & 16.5 & 18.5 \\
Off-topic & 44.4 & 44.1 & 44.2 & 26.6 \\
On-topic & 47.7 & 47.5 & 46.8 & 27.6 \\
Text-grounded & 74.1 & 74.4 & 72.6 & 38.7 \\
\bottomrule
\end{tabular}

%% file: tables/napo_alignment.tex
\begin{tabular}{lrrr}
\toprule
Scenario & Base & NaPO & $\Delta$ \\
\midrule
Multimodal & 69.4 & 67.6 & -1.8 \\
Contradictory & 18.4 & 15.7 & -2.7 \\
Off-topic & 48.4 & 44.4 & -4.0 \\
On-topic & 50.7 & 47.5 & -3.2 \\
Text-grounded & 73.7 & 75.4 & +1.6 \\
\midrule
Overall & 51.3 & 48.9 & -2.5 \\
\bottomrule
\end{tabular}

%% file: tables/vista_related_work_matrix.tex
\begin{table}[t]
\centering
\scriptsize
\setlength{\tabcolsep}{2.8pt}
\renewcommand{\arraystretch}{1.12}
\definecolor{vistaFull}{HTML}{0B7A3B}
\definecolor{vistaPartial}{HTML}{C47A00}
\definecolor{vistaAbsent}{HTML}{8A8A8A}
\newcommand{\full}{\textcolor{vistaFull}{\scalebox{1.16}{\ding{51}}}}
\newcommand{\vistaMarkerLiteral}[1]{%
  \ifdefined\pdfliteral
    \pdfliteral{#1}%
  \else
    \ifdefined\pdfextension
      \pdfextension literal{#1}%
    \fi
  \fi
}
\newcommand{\partialmark}{%
  \begingroup
  \color{vistaPartial}%
  \raisebox{-0.1ex}{%
    \makebox[0.95em][c]{%
      \hbox to 0.8em{%
        \vistaMarkerLiteral{q 0.45 w 3.4 0.4 m 5.057 0.4 6.4 1.743 6.4 3.4 c 6.4 5.057 5.057 6.4 3.4 6.4 c 3.4 0.4 l f 3.4 0.4 m 5.057 0.4 6.4 1.743 6.4 3.4 c 6.4 5.057 5.057 6.4 3.4 6.4 c 1.743 6.4 0.4 5.057 0.4 3.4 c 0.4 1.743 1.743 0.4 3.4 0.4 c S Q}%
        \hss
        \vrule width 0pt height 1.2ex depth 0.15ex%
      }%
    }%
  }%
  \endgroup
}
\newcommand{\absent}{\textcolor{vistaAbsent}{\raisebox{0.55ex}{\rule{0.82em}{0.12em}}}}
\resizebox{\columnwidth}{!}{
\begin{tabular}{lcccccc}
\toprule
\textbf{Benchmark / Dataset} &
\textbf{FP video} &
\textbf{Dialogue} &
\textbf{Decision} &
\textbf{Timing} &
\textbf{Steps} &
\textbf{No-assist} \\
\midrule
PROASSIST & \full & \full & \partialmark & \partialmark & \partialmark & \absent \\
HoloAssist & \full & \full & \partialmark & \partialmark & \partialmark & \partialmark \\
Ego-EXTRA & \full & \full & \partialmark & \partialmark & \partialmark & \partialmark \\
LifeEval & \full & \full & \partialmark & \absent & \partialmark & \partialmark \\
\midrule
ESTP-Bench & \full & \absent & \absent & \full & \absent & \absent \\
ProactiveVideoQA & \partialmark & \absent & \partialmark & \full & \absent & \absent \\
EgoSDQES & \full & \absent & \absent & \full & \absent & \absent \\
OVO-Bench & \absent & \absent & \absent & \partialmark & \absent & \absent \\
StreamingBench & \absent & \absent & \absent & \partialmark & \absent & \absent \\
\midrule
EgoPro-Bench & \full & \partialmark & \partialmark & \partialmark & \partialmark & \partialmark \\
Pro$^{2}$Assist & \full & \partialmark & \full & \partialmark & \full & \partialmark \\
ProAct-75 & \partialmark & \absent & \full & \partialmark & \full & \partialmark \\
TI-PREGO & \full & \absent & \partialmark & \partialmark & \absent & \partialmark \\
MistSense & \full & \absent & \partialmark & \partialmark & \partialmark & \partialmark \\
SafeAgentBench & \absent & \absent & \partialmark & \absent & \partialmark & \full \\
IS-Bench & \absent & \absent & \partialmark & \absent & \partialmark & \full \\
\midrule
\textbf{VISTA} & \textbf{\full} & \textbf{\full} & \textbf{\full} & \textbf{\full} & \textbf{\full} & \textbf{\full} \\
\bottomrule
\end{tabular}
}
\caption{\textbf{Comparison with related benchmarks.}
We compare representative benchmarks along six axes central to VISTA.
\full{}, \partialmark{}, and \absent{} denote primary coverage, partial/adjacent coverage, and no primary coverage, respectively.}
\label{tab:benchmark-comparison}
\end{table}